\documentclass[letterpaper]{article} 
\usepackage{aaai2026}  
\usepackage{times}  
\usepackage{helvet}  
\usepackage{courier}  
\usepackage[hyphens]{url}  
\usepackage{graphicx} 
\usepackage{amssymb}
\usepackage{natbib}  
\usepackage{caption} 
\usepackage{amsmath}
\usepackage{multirow} 
\usepackage{booktabs} 
\usepackage{makecell} 
\usepackage{color,xcolor}         
\usepackage{times}
\usepackage{helvet}
\usepackage{courier}
\usepackage{xcolor}

\usepackage[table, dvipsnames]{xcolor}
\usepackage{algorithm}
\usepackage{algpseudocode}
\usepackage{appendix}

\usepackage{newfloat}
\usepackage{listings}
\definecolor{mygray}{RGB}{210, 210, 210}
\definecolor{mygrayL}{RGB}{240, 240, 240}
\DeclareCaptionStyle{ruled}{labelfont=normalfont,labelsep=colon,strut=off} 
\floatstyle{ruled}
\newfloat{listing}{tb}{lst}{}
\floatname{listing}{Listing}
\nocopyright

\title{Contribution-aware Token Compression for Efficient Video Understanding via Reinforcement Learning}
\author{
    Yinchao Ma, Qiang Zhou, Zhibin Wang, Xianing Chen, Hanqing Yang, Jun Song\thanks{Corresponding Author}, Bo Zheng
}
\affiliations{
    Taobao \& Tmall Group of Alibaba\\
    \{mayinchao.myc, jianchong.zq, jsong.sj\}@taobao.com
    
}

\usepackage{bibentry}

\begin{document}

\maketitle

\begin{abstract}
Video large language models have demonstrated remarkable capabilities in video understanding tasks. 
However, the redundancy of video tokens introduces significant computational overhead during inference, limiting their practical deployment. 
Many compression algorithms are proposed to prioritize retaining features with the highest attention scores to minimize perturbations in attention computations. 
However, the correlation between attention scores and their actual contribution to correct answers remains ambiguous.
To address the above limitation, we propose a novel \textbf{C}ontribution-\textbf{a}ware token \textbf{Co}mpression algorithm for \textbf{VID}eo understanding (\textbf{CaCoVID}) that explicitly optimizes the token selection policy based on the contribution of tokens to correct predictions. 
First, we introduce a reinforcement learning-based framework that optimizes a policy network to select video token combinations with the greatest contribution to correct predictions. 
This paradigm shifts the focus from passive token preservation to active discovery of optimal compressed token combinations. 
Secondly, we propose a combinatorial policy optimization algorithm with online combination space sampling, which dramatically reduces the exploration space for video token combinations and accelerates the convergence speed of policy optimization.
Extensive experiments on diverse video understanding benchmarks demonstrate the effectiveness of CaCoVID.
Codes are available at https://github.com/LivingFutureLab/CaCoVID.
\end{abstract}


\section{Introduction}
Video large language models~\cite{qwenvl-2.5,llava-ov,video-llava} have achieved remarkable advancements in video understanding tasks. 
However, the integration of video data into LLMs introduces substantial computational challenges, primarily due to the numerous processing demands caused by densely encoded video tokens and the inherent quadratic complexity of attention mechanisms.
Thus, compressing video tokens with minimal performance loss has attracted increasing attention for video understanding.

\begin{figure}[t]
    \centering
    \vspace{-2mm}
    \includegraphics[width=0.9\linewidth]{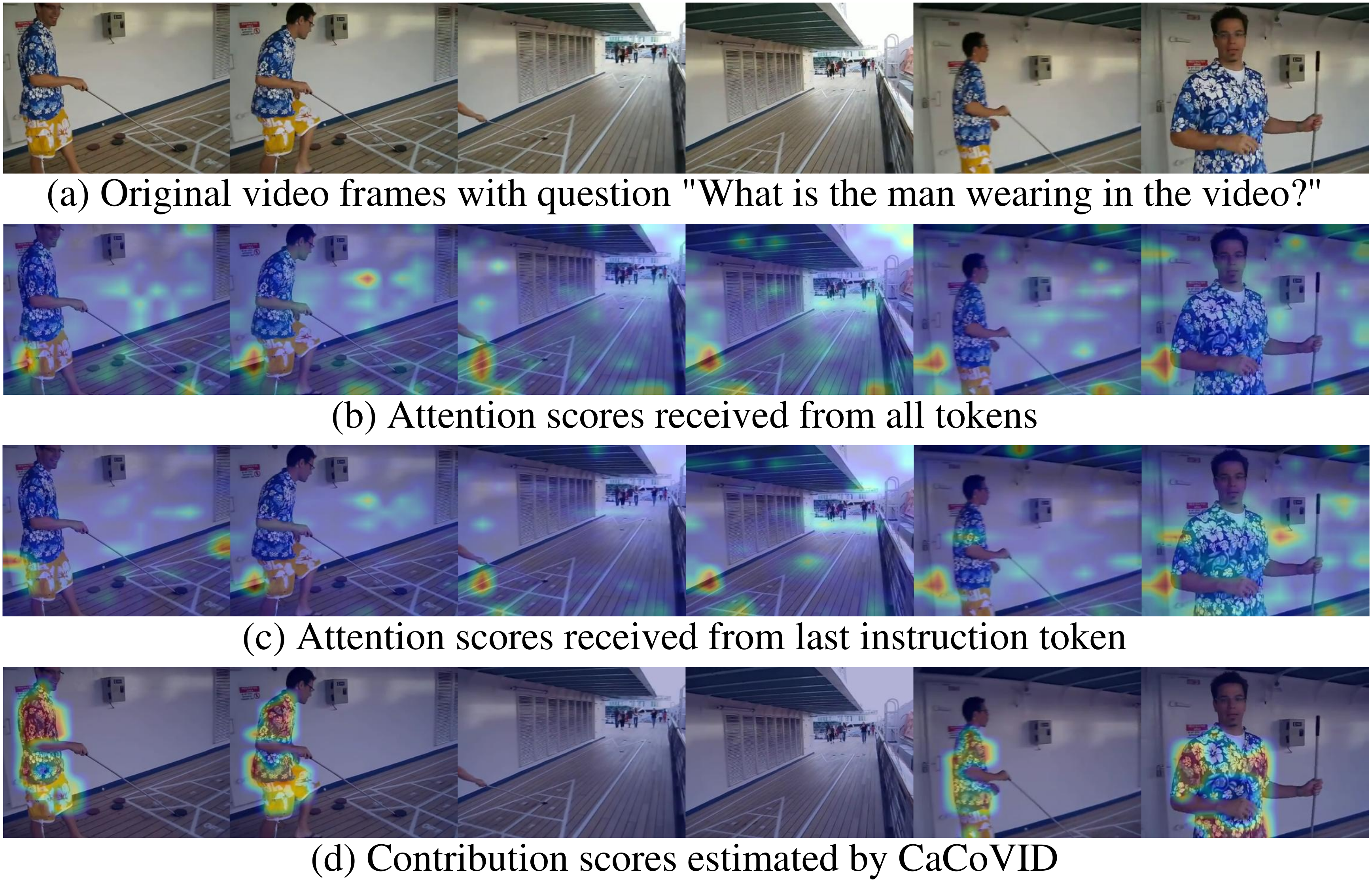}
    \vspace{-2mm}
    \caption{As observed, the attention scores of LLMs demonstrate ambiguity in their correlation with token contributions to correct question answering. 
    Higher attention scores are not allocated to critical tokens such as ``the clothing of the man", which may stem from the visual attention sink phenomenon~\cite{attentionsink}. 
    In contrast, our contribution scores learned through LLM prediction feedback effectively focus on the regions most critical to answering the question correctly - specifically, the ``the clothing of the man" area.}
    \label{fig:intro}
    \vspace{-6mm}
\end{figure}

Numerous studies~\cite{fastv,visionzip} have explored token compression for multimodal large language models (MLLMs). 
The design motivations behind these methods can generally be categorized into two paradigms: content-based token compression and model-based token compression. 
\textbf{Content-based token compression} aims to preserve content diversity or spatio-temporal structure as design objectives, employing handcrafted content metrics to guide token compression.
Typecally, DivPrune~\cite{divprune} prunes features by maximizing the diversity of visual tokens based on the solution of max–min diversity problem (MMDP).
TopV~\cite{topv} comprehensively considers feature similarity, spatial relationships, and central distance to evaluate token importance for pruning.
Although these approaches preserve rich visual contents, their compression strategy is query-agnostic and may prune tokens critical to the query.
\textbf{Model-based token compression} aims to minimize perturbations in LLM inference as design objectives, typically pruning tokens with low attention scores to compress the token sequence.
Typecally, FastV~\cite{fastv} proposes to filter out unimportant tokens by using the average attention score that each token receives from all other tokens as a metric.
PyramidDrop~\cite{pdrop} designs a multi-stage pruning strategy to progressively prune visual tokens at different layers based on the attention scores between the last-instruction token and visual tokens.
However, the correlation between attention scores and their actual contribution to correct answers remains ambiguous, which may lead to suboptimal token compression, as shown in Figure~\ref{fig:intro}(b-c).

To address the above limitations, we shift the focus from passive token preservation to active discovery of optimal compressed token combinations for correct prediction.
To achieve this, there are two critical issues that need to be considered for active exploration of optimal token combinations.
1) \textbf{How to empower active token selection}.
In existing token compression algorithms~\cite{fastv,pdrop}, LLMs fail to actively participate in the token selection process, with tokens instead being passively selected based on handcrafted metrics or pretrained attention scores.
Therefore, the selected tokens may be suboptimal for correct predictions.
To address this, we seek to train a small compression policy network that actively learns to select optimal token combinations with feedback of LLM predictions.
2)~\textbf{How to explore optimal token combinations}.
There currently exists no large-scale video training data annotated with key tokens for token selection optimization.
Although reinforcement learning algorithms~\cite{ppo,grpo} provide solutions to optimize networks via sampling-based exploration with environmental rewards in unsupervised settings, the combinatorial exploration space of $n$ video tokens reaches $2^n$ where $n$ typically exceeds 1000. 
Such astronomically large exploration spaces make native sampling strategies impractical and prone to divergent learning dynamics.
Therefore, it is necessary to develop a new reinforcement learning framework for token combination exploration and optimization.

Based on the above discussions, we propose a novel \textbf{C}ontribution-\textbf{a}ware token \textbf{Co}mpression algorithm for \textbf{VID}eo understanding (\textbf{CaCoVID}) that explicitly optimizes a video token compression policy by actively exploring the contributions of different token combinations to the correct prediction.
Specifically, we design a token compression policy network to estimate the contribution of video tokens and frames to correct prediction by interacting with the query.
The most contributed tokens are retained for LLM inference as shown in Figure~\ref{fig:intro}(d).
Further, to achieve effective combinatorial exploration and policy optimization, we develop a novel combinatorial policy optimization algorithm with online combinatorial space sampling (OCSS).
By partitioning the combinatorial sub-space, OCSS constrains sampling within tokens exhibiting similar contribution scores, thereby dramatically reducing ineffective combinatorial exploration and accelerating the convergence of the policy optimization.
To summarize, the main contributions of this work are:
(1) To the best of our knowledge, we are the first to propose a reinforcement learning-based token compression algorithm (CaCoVID) that ranks and prunes video tokens by directly estimating the contribution to the correct prediction.
(2) We develop a novel combinatorial policy optimization algorithm with online combination space sampling, which dramatically reduces the exploration space for video token combinations and accelerates the convergence of policy optimization.
(3) Extensive experimental results on diverse video understanding benchmarks demonstrate that CaCoVID achieves state-of-the-art performance with lower latency.

\section{Related Work}

In this section, we briefly overview related works on video large language models and visual token compression.
\subsection{Video Large Language Model}
The evolution of video large language models~\cite{minigpt4-video,intern-vl,qwen2-vl} has sparked significant research efforts in developing effective frameworks for spatiotemporal understanding.
Due to the redundancy of video tokens, many studies propose video-specific feature learning frameworks to compress video tokens during LLM training.
Video-ChatGPT~\cite{video-chatgpt} employs temporal and spatial pooling to compress video tokens.
PLLaVA~\cite{pllava} reduces video tokens to the desired feature dimension via adaptive average structure pooling.
Video-XL~\cite{video-xl} utilizes KV sparsification mechanisms in LLMs to compress the visual information into visual summarization tokens.
Further, with the development of unified multimodal frameworks (image, multi-image, video), unified MLLMs attract more attention, $e.g.$ LLaVA-OneVision~\cite{llava-ov} and Qwen2.5-VL~\cite{qwenvl-2.5}, which commonly maintain original frame-level tokens to ensure compatibility across modalities.
However, these approaches still face challenges in video processing due to high token consumption during inference. 
Thus, we propose a contribution-aware token compression algorithm for video understanding, a framework-agnostic and compatible solution.
We optimize a small compression policy network through active token exploration, effectively unlocking the reasoning potential of pretrained models under limited video token budgets without requiring LLMs retraining.

\subsection{Visual Token Compression}
Visual feature compression~\cite{tome,tempme} has gained increasing attention due to the computational burdens caused by redundant visual tokens in multimodal large language models (MLLMs). 
The design motivations behind these methods can generally be categorized into two paradigms: content-based token compression and model-based token compression.

\noindent
\textbf{Content-based token compression} aims to preserve content diversity or spatio-temporal structure as design objectives, employing handcrafted content metrics to guide token compression. 
TopV~\cite{topv} optimizes token pruning through solving a cost matrix based on token distance and similarity.
DivPrune~\cite{divprune} enhances diversity preservation by solving a max-min diversity problem to retain structurally distinct tokens. 
VisionZip~\cite{visionzip} reduces spatial redundancy through attention-driven feature analysis in vision encoders.
PruneVID~\cite{prunevid} merges static tokens in video segments and reduces spatial redundancy via DPC-KNN~\cite{dpc-knn} merging.
Although preserving diverse visual contents, their compression strategies are query-agnostic and may prune tokens critical to queries.

\begin{figure*}[t]
    \centering
    \includegraphics[width=0.8\linewidth]{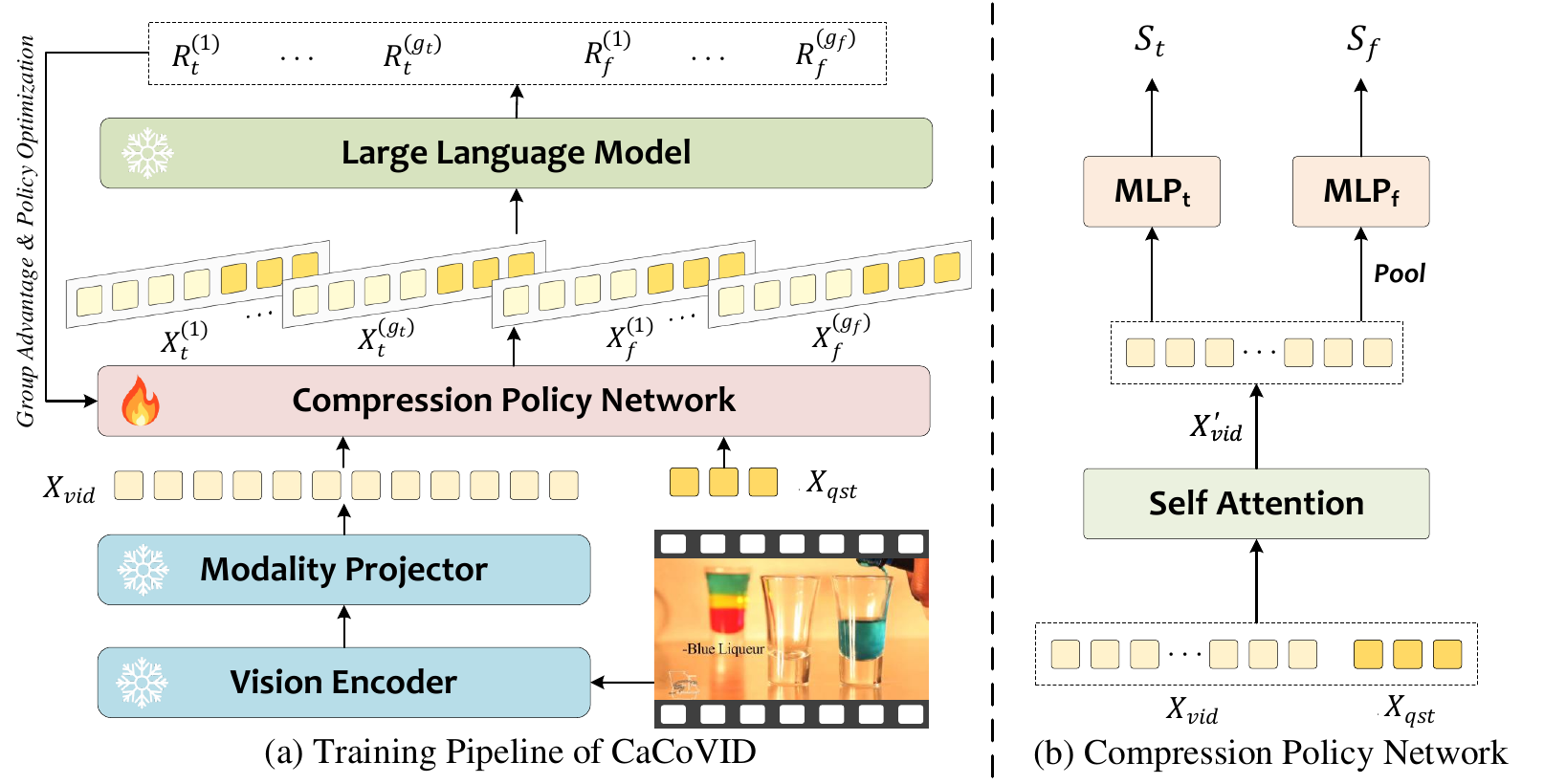}
    \caption{(a) illustrates the training pipeline of contribution-aware token compression for video understanding (CaCoVID), which optimizes the compression policy network by exploring the contributions of different token combinations to correct predictions. (b) presents the architecture of the compression policy network, which builds interactions between video tokens and question tokens in a self-attention mechanism and estimates the contributions of video tokens or frames to the correct prediction by MLPs.}
    \label{fig:arch}
\end{figure*}

\noindent
\textbf{Model-based token compression} aims to minimize perturbations in model inference as design objectives, typically pruning tokens with low attention scores to compress the token sequence. 
FastV~\cite{fastv} prunes visual tokens in early LLM layers through average attention scores received from other tokens.
PyramidDrop~\cite{pdrop} implements multi-stage pruning by progressively eliminating low-attention tokens at different network depths. 
DyCoke~\cite{dycoke} dynamically prunes key-value cache via prioritizing tokens with high attention scores to the predicted token.
SparseVLM~\cite{sparsevlm} filters key visual tokens through attention scores with question tokens.
However, the correlation between attention scores and their actual contribution to correct answers remains ambiguous, which may lead to suboptimal token compression.

To address the above limitations, we shift the focus from passive token preservation to active discovery of optimal compressed token combinations for correct prediction.
We achieve this by exploring the contribution of different token combinations to the correct prediction and optimizing a policy network that ranks and prunes video tokens without requiring retraining of the LLM.

\section{Method}

In this section, we first present the preliminaries of video large language models (Video LLMs), elaborating on the critical importance of video token compression for Video LLMs. 
Subsequently, we formally introduce our proposed contribution-aware token compression for video understanding, which incorporates a learnable compression policy network and a combinatorial policy optimization algorithm.

\subsection{Preliminaries}
\noindent
\textbf{Video Large Language Models}. Video LLMs typically encode densely sampled video frame features through a visual encoder, then map these visual features into the LLM's embedding space via a modality projector, enabling alignment between video tokens and language tokens. The concatenation of video tokens $X_{vid} \in \mathbb{R}^{n_{vid} \times d}$, and question tokens $X_{qst} \in \mathbb{R}^{n_{qst} \times d}$ serves as the input to LLM for the prefilling and decoding.

\noindent
\textbf{Computation Complexity}. The encoder of LLM typically consists of the self-attention mechanism and the feed-forward network (FFN). 
The total computational costs of LLM can be expressed as ${FLOPs} = T \times (4nd^2 + 2n^2d + 2ndm)$.
Here, $T$ is the number of transformer layers, $n=n_{vis} + n_{qst}$ is the input sequence length, $d$ is the hidden dimension size, and $m$ denotes the intermediate size of the FFN.
The computational complexity quadratically depends on the sequence length $n$.
In video LLMs, video token length $n_{vis}$ dominates the total ($e.g.$, $32\times196=6,272$ video tokens in LLaVA-OneVision), while the question token length $n_{qst}$ typically remain below 512. Consequently, over 90\% of FLOPs arise from attention interactions with video tokens.
This highlights urgent demand for video token compression to eliminate redundant spatial-temporal information.

\subsection{Contribution-aware Token Compression}
\label{sec:3.2}
In existing token compression algorithms~\cite{fastv,pdrop}, LLMs fail to actively participate in the token selection process, with tokens instead being passively selected based on handcrafted metrics or pretrained attention scores.
The compression objectives may not align with the goals of correct prediction. 
To address this, we seek to actively explore critical tokens through feedback from LLM and optimize a small network to estimate the contribution of video tokens to correct predictions for token compression.

\noindent
\textbf{Compression Policy Network}.
The compression policy network is composed of self-attention mechanisms and two multi-layer perceptrons (${\rm MLP_t}$ and ${\rm MLP_f}$), which output two-dimensional logits, as shown in Figure~\ref{fig:arch}(b).
Given encoded video tokens $X_{vid} \in \mathbb{R}^{n_{vid} \times d}$ and question tokens $X_{qst} \in \mathbb{R}^{n_{qst} \times d}$, the self-attention mechanism first establishes cross-modal interactions between video tokens and text tokens to generate question-aware video tokens $X^\prime_{vid}$. 
Here, $n_{vid}=t\times h\times w$, and $t,h,w$ are the number of frames, height and width of encoded video tokens.
Subsequently, these video tokens are fed into two MLPs to estimate the contribution of each video token and frame to correctly answering the question.
Formally,
\begin{gather}[X_{vid}^{\prime},X_{qst}^{\prime}]={\rm SelfAttn}([X_{vid},X_{qst}]),\\
S_f = {\rm MLP_f}({\rm Pool}(X_{vid}^{\prime}))\in \mathbb{R}^{t\times 2},\\
S_t = {\rm MLP_t}(X_{vid}^{\prime})\in \mathbb{R}^{n_{vid}\times 2}, \\
\hat{S}_t=S_t[:,1]-S_t[:,0],\hat{S}_f=S_f[:,1]-S_f[:,0],
\end{gather}
where ${\rm Pool(\cdot)}$ means pooling along the spatial dimension, 
the two output channels of MLPs indicate whether the corresponding token or frame should be selected for answering the question.
The difference between the two channels can be regarded as the potential contribution of tokens $\hat{S}_t$ or frames $\hat{S}_f$ to correctly answering the question.
During inference, $\hat{S}_t$ is used to select the most contributed token in each frame and $\hat{S}_f$ is used to allocate token budget across frames.

\noindent
\textbf{Combinatorial Policy Optimization}.
There currently exists no large-scale video training data annotated with key tokens for token selection optimization, which precludes the direct application of supervised learning in policy network training.
Traditional reinforcement learning algorithms~\cite{ppo,grpo} for large language models typically perform exploration in an autoregressive manner over the vocabulary-sized dimension, and are usually built upon pretrained models. 
As a result, the exploration space is relatively constrained, which facilitates model convergence.
However, for thousands of video token selections, autoregressive estimation of individual token contributions is highly inefficient. 
Meanwhile, parallel prediction of all token contributions entails an exponentially large combinatorial exploration space ${\rm O}(2^n)$, rendering native sampling-based exploration strategies computationally infeasible and prone to divergent learning dynamics.

\begin{figure}
    \centering
    \includegraphics[width=\linewidth]{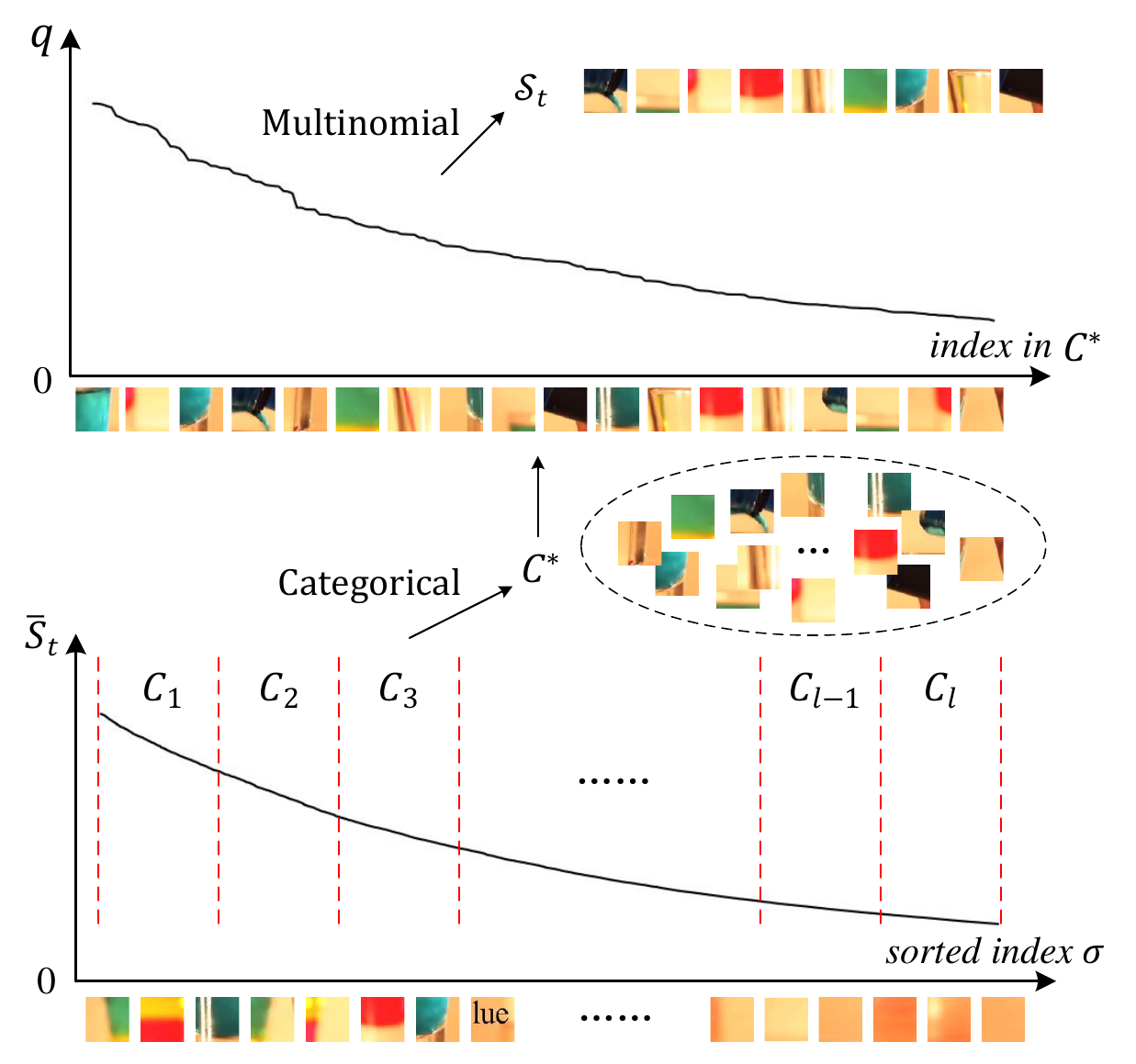}
    \caption{Illustration of the online combinatorial space sampling strategy. Tokens are first sorted by their contribution scores estimated by the policy network and partitioned into combinatorial sub-spaces.
    Then, a categorical distribution is applied to sample the subspace according to the sum of contribution scores within each subspace. 
    Finally, tokens within the selected subspace are sampled according to a multinomial distribution to acquire the final token combination.}
    \label{fig:sample}
\end{figure}

How to \textit{narrow down the exploration space} and \textit{leverage online learning experiences} to minimize ineffective exploration constitutes critical issues in policy optimization.
To address the above issues, we propose a combinatorial policy optimization algorithm (CPO), integrating a novel online combinatorial space sampling strategy.
The goal of the compression policy network is to discover the video token combinations that contribute most to correct predictions, which drives our focus toward systematically exploring token combinations with the highest potential to guide the LLM toward correct answers for given questions, rather than indiscriminately exploring arbitrary combinations.
To this end, we develop the online combinatorial space sampling strategy (OCSS), which divides the token combinatorial space into multiple combinatorial sub-spaces according to the estimated contribution scores, such that tokens with similar contribution levels are more likely to be sampled together, as shown in Figure~\ref{fig:sample}.
Given a token sampling ratio $ r $, we first sort all tokens based on their estimated contribution scores of the policy network, and then divide them into $ l=1/(\lambda\times r) $ combinatorial sub-spaces, each containing $ m=\lambda \times r \times n_{vid} $ tokens, $\lambda=2$ by default.
Formally,
\begin{gather}
    \sigma = \text{argsort}(\hat{S}_t),\\
    C_i = \left\{ \sigma_j \mid j \in \left[ (i-1)m + 1,\, im \right] \right\}, i = 1, 2, \dots, l
\end{gather}
where $ \sigma $ denotes the sorted index arrangement of video token, $ C_i \subset \mathcal{Z}_t $ is the token index set of $ i^{th} $ combinatorial sub-space, $\mathcal{Z}_t=\{1, 2, \dots, n_{{vid}}\}$ is the universal set of the video token indexes.
Then, we employ probabilistic sampling based on the total contribution scores of tokens in each combinatorial sub-space.
Formally,
\begin{gather}
    \bar{S}_t = Softmax(\hat{S}_t)\in \mathbb{R}^{n_{vid}},\\
    p_i = \sum_{k \in C_i} \bar{S}_t[k],\\
    C^* \sim \text{Categorical}([p_1, p_2, \dots, p_l]),
\end{gather}
where $ C^* $ is the sampled combinatorial sub-space, ${\rm Categorical(\cdot)}$ denotes the categorical distribution.
Subsequently, a probabilistic sampling is conducted within the selected combinatorial sub-space to further explore its internal token combinations.
Formally,
\begin{gather}
    q_k = \frac{\exp(\hat{S}_t[k])}{\sum_{u \in C^*} \exp(\hat{S}_t[u])}, \quad k \in C^*,\\
    \mathcal{S}_t \sim \text{Multinomial}\left(r \cdot n_{vid},\, [q_k]_{k \in C^*}\right), \\
    \hat{X}_{t} = X_{vid}[\mathcal{S}_t,:]\in\mathbb{R}^{(r\cdot n_{vid})\times d},
\end{gather}
where $ \mathcal{S}_t \subset C^* $ is the final sampled $ r \cdot n_{\text{vid}} $ video token index, ${\rm Multinomial(\cdot,\cdot)}$ denotes the multinomial distribution, $\hat{X}_{t}$ denotes the sampled video tokens.

During each training iteration, we sample $g_t$ groups of video tokens $\{\hat{X}_{t}^{(i)}\}_{i=1}^{g_t}$ and feed them into the LLM along with question tokens $X_{qst}$ for reasoning, as shown in Figure~\ref{fig:arch}(a).
We evaluate the correctness of the predictions by comparing them with the answer $\mathcal{A}$ as the reward for the video token combinations. Formally,
\begin{gather}
    X^{(i)}_t=[\hat{X}_{t}^{(i)},X_{qst}], \quad i = 1, 2, \dots, g_t\\
    {Y}^{(i)}_t = \text{LLM}(X^{(i)}_t), 
    R^{(i)}_t={\rm Reward}({Y}^{(i)}_t,\mathcal{A}),
\end{gather}
where the reward functions ${\rm Reward}(\cdot,\cdot)$ are consistent with Video-R1~\cite{video-r1}. Then, we compute the group advantage for policy optimization.
Formally,
\begin{gather}
    A^{(i)}_t = \frac{R^{(i)}_t-{\rm mean}(\{R^{(i)}_t\}_{i=1}^{g_t})}{{\rm std}(\{R^{(i)}_t\}_{i=1}^{g_t})}.
\end{gather}
Finally, we optimize the policy model for token contribution estimation by maximizing the following objective,
\begin{gather}
    \mathcal{J}_{CPO}^t(\theta)=\mathbb{E}_{j\in\mathcal{Z}_t}\left[\frac{1}{g_t}\sum_{i=1}^{g_t}{r_{i,j}^{clip}(\theta)}\right],\\
    \text{\footnotesize $r_{i,j}^{clip}(\theta)={\rm min}\left[r_{i,j}(\theta)A^{(i)}_t,{\rm clip}\left(r_{i,j}(\theta),1-\epsilon_{low},1+\epsilon_{high}\right)A^{(i)}_t\right]$}, \\
    r_{i,j}(\theta)=\frac{\pi_{\theta}^t\left(j|[X_{vid},X_{qst}],\mathcal{S}_t^{(i)}\right)}{\pi_{\theta_{old}}^t\left(j|[X_{vid},X_{qst}],\mathcal{S}_t^{(i)}\right)}, \quad j\in \mathcal{Z}_t
\end{gather}
where $\epsilon_{low},\epsilon_{high}$ is a clipping hyper-parameter~\cite{dapo} for stabilizing training, $\pi_{\theta}^t\left(\cdot|\cdot,\cdot\right)$ and $\pi_{\theta_{old}}^t\left(\cdot|\cdot,\cdot\right)$ denote the current and old compression policy network with parameters $\theta$ and $\theta_{old}$, whose output can be expressed as,
    \[
\pi_{\theta}^t\left(j|[X_{vid},X_{qst}],\mathcal{S}_t^{(i)}\right) = 
\begin{cases} 
\sigma(S_t)[j,1], & \text{if } j\in \mathcal{S}_t^{(i)} \\
\sigma(S_t)[j,0], & \text{if } j\in \mathcal{Z}_t\setminus\mathcal{S}_t^{(i)}
\end{cases}
\]
where $\sigma(\cdot)$ means softmax along the channel dimension.
Similarly, we sample $g_f$ groups of video frames through the online combinatorial space sampling strategy and calculate the group advantage function through the prediction of LLM to optimize the compression policy network for video frames, as detailed in the appendix.

\noindent
\textbf{Data Exploration Efficiency}.
The effective utilization of training samples during model training is another critical aspect for policy network optimization, which involves three key considerations.
1) \textit{Ineffective sample filtering}. Certain video questions can be answered correctly through blind testing (without video input), rendering these samples non-informative for policy network learning. 
We address this by filtering out simple samples through blind testing. 
2) \textit{Experience replay}. To fully explore the video token combinations, we iterate each training sample multiple times combined with experience replay mechanisms to generate more exploration experience. 
3) \textit{Dynamic sample ratio}. Specific video-question pairs may consistently produce correct/incorrect responses under the sample ratio $r$, which hinders effective exploration.
We introduce a dynamic sample ratio strategy during training. 
For multiple iterations of a training sample, if the average reward from the previous iteration exceeds $\alpha_{high}$, we decrease the video sample ratio by half in the next iteration; if the average reward falls below $\alpha_{low}$, we increase the video sample ratio by two times.
The detailed training algorithm is shown in the appendix.

\section{Experiment}

\subsection{Experimental Setting}

\begin{table*}[t!]
\centering
\vspace{-5mm}
\resizebox{0.83\linewidth}{!}
{
\begin{tabular}{c|c|c|cc|cccc|cc}
\toprule
 \multirow{3}{*}{Method} & \multirow{3}{*}{\makecell{Equivalent\\Retention Ratio}}  & \multirow{2}{*}{\makecell{LongVideoBench}} & \multicolumn{2}{c|}{MLVU} & \multicolumn{4}{c}{VideoMME}                              & \multicolumn{2}{|c}{Avg. Acc.} \\ \cline{4-11}
                                                       &   &                                                          &            Dev                     &Test                       & Short       & Medium       & Long   & Overall             & \multirow{2}{*}{Score}       & \multirow{2}{*}{\%}         \\
                                                                      &      & \textit{1$\sim$60min}                    & \multicolumn{2}{c|}{\textit{3$\sim$120min}}        & \textit{1$\sim$3min} & \textit{3$\sim$30min} & \textit{30$\sim$60min} &  
                                                                      \textit{1$\sim$60min} &&            \\ \midrule
LLaVA-OneVision-7B      & 100\%      & 56.4 & 63.0 & 47.7 & 70.3 & 56.8 & 48.6 & 58.6 & 56.4 & 100 \\ 
\midrule
FastV~\cite{fastv}         & 25\%    & 53.5 & 59.0 & 41.7 & 64.1 & 53.1 & 48.1 & 55.1 & 52.3 & 92.7 \\
VisionZip~\cite{visionzip} & 25\%    & 56.0 & 61.5 & 43.5 & 69.0 & 55.0 & 48.3 & 57.4 & 54.6 & 96.8 \\
DivPrune~\cite{divprune}   & 25\%    & \textbf{56.4} & \textbf{62.4} & 44.1 & 69.0 & 54.6 & 47.9 & 57.1 & 55.0 & 97.5 \\
PruneVID~\cite{prunevid}   & 25\%    & 56.1 & 60.9 & 45.4 & 68.6 & \textbf{56.7} & \textbf{48.9} & 58.0 & 55.1 & 97.7 \\
FrameFusion~\cite{framefusion}& 25\% & 55.7 & 61.1 & 44.2 & 68.7 & 55.2 & 48.2 & 57.4 & 54.6 & 96.7 \\
CaCoVID                    & 25\%    & 56.2 & 62.3 & \textbf{46.3} & \textbf{70.7} & 56.3 & 48.4 & \textbf{58.5} & \textbf{55.8} & \textbf{98.9} \\
\midrule
FastV~\cite{fastv}         & 20\%    & 53.0 & 58.3 & 42.0 & 62.1 & 52.2 & 46.9 & 53.7 & 51.8 & 91.8 \\
VisionZip~\cite{visionzip} & 20\%    & 55.7 & 61.5 & 44.7 & 68.8 & \textbf{55.9} & 47.8 & \textbf{57.5} & 54.9 & 97.2 \\
DivPrune~\cite{divprune}   & 20\%    & 55.6 & \textbf{62.2} & 43.9 & 68.4 & 54.4 & 48.1 & 57.1 & 54.7 & 96.9 \\
PruneVID~\cite{prunevid}   & 20\%    & 54.9 & 61.4 & \textbf{45.7} & 67.4 & 54.2 & 47.8 & 56.8 & 54.7 & 96.9 \\
FrameFusion~\cite{framefusion}& 20\% & 55.1 & 60.5 & 44.2 & 67.7 & 54.4 & 47.2 & 56.4 & 54.1 & 95.8 \\
CaCoVID                    & 20\%    & \textbf{56.5} & 62.0 & 45.3 & \textbf{69.7} & 54.3 & \textbf{48.6} & \textbf{57.5} & \textbf{55.3} & \textbf{98.1} \\
\midrule
FastV~\cite{fastv}         & 15\%    & 49.8 & 55.5 & 41.7 & 58.0 & 52.7 & 46.0 & 52.2 & 49.8 & 88.3 \\
VisionZip~\cite{visionzip} & 15\%    & 54.0 & 61.2 & 41.9 & 65.7 & 53.4 & \textbf{48.2} & 55.8 & 53.2 & 94.3 \\
DivPrune~\cite{divprune}   & 15\%    & 55.1 & \textbf{61.9} & 43.9 & 66.6 & \textbf{54.9} & 48.0 & 56.5 & 54.3 & 96.3 \\
PruneVID~\cite{prunevid}   & 15\%    & 55.1 & 59.9 & 44.3 & 66.6 & 54.4 & 47.9 & 56.3 & 53.9 & 95.5 \\
FrameFusion~\cite{framefusion}& 15\% & 52.9 & 58.4 & 41.8 & 66.2 & 53.0 & 46.6 & 55.3 & 52.1 & 92.3 \\
CaCoVID                    & 15\%    & \textbf{56.8} & 61.2 & \textbf{44.9} & \textbf{69.3} & 54.4 & 47.4 & \textbf{57.1} & \textbf{55.0} & \textbf{97.5} \\
\midrule
FastV~\cite{fastv}         & 10\%    & 47.6 & 52.9 & 35.2 & 53.3 & 48.0 & 44.2 & 48.5 & 46.1 & 81.6 \\
VisionZip~\cite{visionzip} & 10\%    & 48.9 & 57.6 & 41.5 & 58.6 & 52.3 & 46.2 & 52.4 & 50.1 & 88.8 \\
DivPrune~\cite{divprune}   & 10\%    & 53.3 & \textbf{60.7} & \textbf{44.9} & 63.9 & 52.9 & 46.2 & 54.3 & 53.3 & 94.5 \\
PruneVID~\cite{prunevid}   & 10\%    & 54.4 & 59.7 & 42.7 & 66.0 & 52.4 & 46.3 & 54.9 & 52.9 & 93.8 \\
FrameFusion~\cite{framefusion}& 10\% & 49.1 & 56.5 & 39.8 & 59.6 & 49.8 & 45.1 & 51.5 & 49.2 & 87.2 \\
CaCoVID                    & 10\%    & \textbf{55.2} & 60.6 & 44.6 & \textbf{67.6} & \textbf{54.9} & \textbf{46.6} & \textbf{56.3} & \textbf{54.2} & \textbf{96.1} \\
\midrule\midrule
Qwen2.5-VL-3B           & 100\%      & 54.3 & 62.1 & 45.6 & 71.4 & 60.1 & 49.6 & 60.4 & 55.6 & 100 \\ 
\midrule
FastV~\cite{fastv}         & 25\%    & 49.5 & 57.0 & 43.0 & 66.0 & 56.0 & 48.2 & 56.7 & 51.6 & 92.7 \\
VisionZip~\cite{visionzip} & 25\%    & 51.7 & 58.5 & 43.8 & 67.9 & 55.6 & {48.9} & 57.4 & 52.9 & 95.1 \\
DivPrune~\cite{divprune}   & 25\%    & \textbf{53.4} & 58.8 & \textbf{46.4} & {68.4} & 55.2 & 47.9 & 57.2 & 53.9 & 97.0 \\
PruneVID~\cite{prunevid}   & 25\%    & 53.0 & 59.4 & 43.0 & 64.8 & 53.1 & 47.7 & 55.2 & 52.6 & 94.6 \\
FrameFusion~\cite{framefusion}& 25\% & 52.4 & 58.5 & 44.8 & \textbf{69.1} & \textbf{57.8} & \textbf{50.0} & \textbf{59.0} & 53.7 & 96.5 \\
CaCoVID                    & 25\%    & 53.3 & \textbf{59.9} & 46.2 & 67.7 & {56.1} & 48.8 & {57.5} & \textbf{54.2} & \textbf{97.5} \\
\midrule
FastV~\cite{fastv}         & 20\%    & 48.2 & 55.6 & 41.3 & 64.1 & 54.4 & 47.3 & 55.3 & 50.1 & 90.1 \\
VisionZip~\cite{visionzip} & 20\%    & 51.9 & 57.9 & 43.5 & 66.3 & 54.6 & 48.2 & 56.4 & 52.4 & 94.2 \\
DivPrune~\cite{divprune}   & 20\%    & 52.3 & 58.4 & \textbf{45.6} & 67.6 & 53.3 & 47.2 & 56.0 & 53.1 & 95.5 \\
PruneVID~\cite{prunevid}   & 20\%    & 52.1 & 58.1 & 43.7 & 64.1 & 53.4 & 46.9 & 54.8 & 52.2 & 93.8 \\
FrameFusion~\cite{framefusion}& 20\% & 50.7 & 57.3 & 43.4 & \textbf{68.0} & \textbf{56.6} & \textbf{49.6} & \textbf{58.0} & 52.3 & 94.1 \\
CaCoVID                    & 20\%    & \textbf{52.8} & \textbf{60.2} & 44.8 & {67.9} & \textbf{56.6} & {48.4} & {57.6} & \textbf{53.9} & \textbf{96.9} \\
\midrule
FastV~\cite{fastv}         & 15\%    & 47.8 & 53.7 & 37.3 & 61.3 & 52.6 & 48.2 & 54.0 & 48.2 & 86.7 \\
VisionZip~\cite{visionzip} & 15\%    & 52.0 & 57.7 & \textbf{43.8} & {66.6} & 53.8 & {48.6} & {56.3} & 52.4 & 94.3 \\
DivPrune~\cite{divprune}   & 15\%    & 51.5 & 57.7 & 42.0 & 66.3 & 52.4 & 48.0 & 55.6 & 51.7 & 92.9 \\
PruneVID~\cite{prunevid}   & 15\%    & 52.1 & 58.0 & 42.1 & 61.8 & 51.7 & 47.4 & 53.6 & 51.5 & 92.5 \\
FrameFusion~\cite{framefusion}& 15\% & 49.9 & 55.4 & 41.5 & \textbf{67.0} & \textbf{55.0} & \textbf{49.2} & \textbf{57.1} & 51.0 & 91.6 \\
CaCoVID                    & 15\%    & \textbf{52.4} & \textbf{58.2} & 43.2 & 66.2 & {54.4} & 47.8 & 56.1 & \textbf{52.5} & \textbf{94.4} \\
\midrule
FastV~\cite{fastv}         & 10\%    & 45.0 & 50.8 & 32.0 & 55.2 & 48.1 & 46.1 & 49.8 & 44.4 & 79.9 \\
VisionZip~\cite{visionzip} & 10\%    & \textbf{51.5} & 57.2 & 42.4 & 63.8 & 52.1 & 46.6 & 54.2 & 51.3 & 92.3 \\
DivPrune~\cite{divprune}   & 10\%    & 50.5 & 56.4 & 40.8 & \textbf{64.3} & 51.4 & 47.2 & \textbf{54.3} & 50.5 & 90.9 \\
PruneVID~\cite{prunevid}   & 10\%    & 51.3 & 56.6 & 40.4 & 59.7 & 51.1 & 46.6 & 52.4 & 50.2 & 90.2 \\
CaCoVID                    & 10\%    & \textbf{51.5} & \textbf{57.9} & \textbf{44.2} & 63.1 & \textbf{52.2} & \textbf{47.6} & \textbf{54.3} & \textbf{52.0} & \textbf{93.4} \\
\bottomrule
\end{tabular}
}
\vspace{-1mm}
\caption{
Comparison of state-of-the-art methods on LLaVA-OneVision-7B and Qwen2.5-VL-3B. 
For compression algorithms such as FastV and FrameFusion applied in the LLM prefilling phase, we report the results with layer-averaged retention ratio termed equivalent retention ratio as FrameFusion does.
Bold highlights optimal performance with similar retention ratios.}
\label{tab:sota_ov}
\vspace{-3mm}
\end{table*}

\noindent
\textbf{Training Data}.
Our training data is sourced from Video-R1, which aggregates over 116,000 video data. 
To ensure reliable reward estimation, we specifically selected multiple-choice QA pairs as our training corpus, comprising 104,000 instances.
Further, as detailed in Section~3.2, we conducted blind evaluations on these multiple-choice QA data using LLaVA-OneVision-7B~\cite{llava-ov} and Qwen2.5-VL-3B~\cite{qwenvl-2.5} to filter out trivial questions that could be correctly answered without video context, ultimately retaining a refined dataset of 31,000 high-quality video-relevant questions for policy network training.

\noindent
\textbf{Training Details}.
We implement the contribution-aware token compression algorithm for video understanding (CaCoVID) on LLaVA-OneVision-7B and Qwen2.5-VL-3B. 
The parameters of the vision encoder, projector, and large language model are frozen during training, with only the compression policy network being optimized. 
The attention mechanism of the compression policy network is initialized using pretrained parameters from the first layer of the LLM, while the MLPs are initialized via Xavier init~\cite{xavier}. 
We employ distinct learning rates of $1\times 10^{-7}$ for the attention mechanism and $1\times 10^{-6}$ for MLPs. 
The token sample ratio is set to $r=0.02$ and group number is set to $g_t=24,g_f=8$ during training, with each sample undergoing 5 iterative optimization steps with experience replay, as discussed in Section~3.2.
We set $\alpha_{high}=0.875,\alpha_{low}=0.125,\epsilon_{high}=0.28,\epsilon_{low}=0.2$. 
The policy network parameters are optimized through our CPO, completing full training in a single epoch.
The training process takes about 10 hours on 8$\times$H100 GPUs with one batch per GPU.
The random seed for training is set as $42$.

\noindent
\textbf{Benchmarks}.
We evaluate CaCoVID on diverse video understanding benchmarks including LongVideoBench~\cite{longvideobench}, MLVU~\cite{mlvu}, and VideoMME~\cite{videomme}. 
These benchmarks consist of videos with varying temporal scales, task complexities, and domain diversities, providing a comprehensive and generalized metric framework for evaluating performance.

\noindent
\textbf{Inference Details}.
We reserve the top contribution tokens of each frame estimated by the compression policy network to answer the given question.
The number of tokens $n_j$ reserved in $j^{th}$ frame is determined by the estimated contribution of the frame $n_j={\rm Softmax}(\hat{S}_f)[j]\cdot r\cdot n_{vid}$.
To preserve the spatio-temporal structure of videos, we sample tokens uniformly in the video as complementary to the top-contributed tokens, with these spatio-temporal tokens accounting for 50\% of the retention token allocation.

\vspace{-2mm}
\subsection{State-of-the-art Comparisons}
We comprehensively evaluate our CaCoVID under various retention ratios $r\in \{10\%,15\%,20\%,25\%\}$ and compare it with state-of-the-art methods.
The implementation details of these algorithms being compared are provided in the appendix.
Accuracy is the main evaluation metric.

\noindent
\textbf{Results on LLaVA-OneVision and Qwen2.5-VL.}
During evaluation on LLaVA-OneVision-7B, we uniformly sample 32 frames from the video, and each frame is encoded into 196 tokens.
During evaluation on Qwen2.5-VL-3B, we uniformly sample 64 frames from the video, and each frame is encoded into no more than 256 tokens.
As shown in Table~\ref{tab:sota_ov}, we compare the performance of different compression algorithms under identical hardware and software environments. 
Our method consistently outperforms the previous state-of-the-art approach across various retention ratios. 
Compared to content-based compression algorithms like VisionZip, DivPrune, and PruneVID, CaCoVID achieves question-aware token contribution estimation, effectively retaining tokens critical for answering specific questions. 
Compared to model-based compression algorithms like FastV and FrameFusion, CaCoVID aligns the optimization objectives of compression algorithms with the prediction accuracy of LLMs, thereby enhancing the overall performance.
\begin{table}[t]
    \centering
    \resizebox{\linewidth}{!}
    {
        \begin{tabular}{c|c|c|c}
        \toprule
            Method   & Compression Time (ms)$\downarrow$ & LLM Prefilling Time (ms)$\downarrow$ & Avg. Acc.$\uparrow$ \\ 
            \midrule
            Vanilla  & -     & 200.4 & 56.4 (100\%) \\ \midrule
            DivPrune & 134.3 & 53.2 & 55.0 (97.5\%) \\ 
            PruneVID & 34.1  & 53.2 & 55.1 (97.7\%) \\\midrule
            CaCoVID  & 11.2  & 53.2 & 55.8 (98.9\%) \\
        \hline
        \end{tabular}
    }
    \caption{Comparison of compression efficiency on LLaVA-OneVision-7B with 25\% retention ratio.}
    \vspace{-3mm}
    \label{tab:efficiency}
\end{table}

\noindent
\textbf{Comparison of compression efficiency.}
As shown in Table~\ref{tab:efficiency}, CaCoVID achieves superior compression speed compared to previous compression algorithms, while demonstrating significantly improved performance. 
On one hand, the policy network in CaCoVID can estimate the contribution of each token to accurate predictions in parallel, thereby reducing compression latency. 
On the other hand, our contribution-aware token compression for video understanding explicitly aligns the objective of the compression policy with the correct prediction of LLM, thereby achieving superior performance.

\subsection{Ablation Study}
We perform ablation study on LLaVA-OneVision-7B with retention ratio 25\%.

\begin{table}[t]
    \centering
    \resizebox{0.8\linewidth}{!}
    {
        \begin{tabular}{c|c|c|cc}
        \toprule
            Sample       & Range & $r\%$ & MLVU$_{dev}$ & VideoMME  \\ \midrule
            OCSS         & Video & 2\%  & 61.7 & 57.2  \\
            OCSS         & Frame & 2\%  & 62.3 & 58.5  \\ \midrule
            Random       & Frame & 2\%  & 57.9 & 54.4  \\
            Multinomial  & Frame & 2\%  & 59.2 & 55.1  \\
            OCSS         & Frame & 1\%  & 62.2 & 58.3  \\
            OCSS         & Frame & 2\%  & 62.3 & 58.5  \\
            OCSS         & Frame & 5\%  & 61.6 & 57.3  \\
        \hline
        \end{tabular}
    }
    \caption{Performance with different online sampling strategy during training.}
    \vspace{-5mm}
    \label{tab:sample}
\end{table}

\noindent
\textbf{Analysis of the online combinatorial space sampling}.
Firstly, given the token sampling ratio of $r\%$, we explored two training strategies: concatenating $r\%$ tokens uniformly sampled from each frame by OCSS (Frame) versus directly sampling $r\%$ tokens from the entire video token set by OCSS (Video). 
As shown in Table~\ref{tab:sample}, we find that intra-frame token sampling followed by concatenation yields superior performance. 
The underlying reason could be that the combinatorial space of intra-frame sampling reduces sampling complexity while preserving the complete temporal structure of the video, which facilitates better understanding of video content by the LLM. 
Furthermore, we compare different sampling strategies including random sampling, multinomial probability sampling, and our proposed online combinatorial space sampling (OCSS). 
Experimental results demonstrate that OCSS significantly outperforms both random sampling and multinomial distribution sampling. 
This improvement stems from OCSS's ability to narrow down the exploration space and fully leverage online learning experiences to minimize inefficient exploration, thereby enabling the compression policy network to efficiently discover optimal token combinations for question answering.
Furthermore, we can observe that training with smaller sampling ratios tends to yield higher performance. 
This is because numerous token combinations can provide correct answers to the question under high sampling ratios, leading to ineffective policy optimization.
Meanwhile, excessively small sampling ratios could also hinder the sampling of effective token combinations required for answering questions, thereby slightly reducing performance.

\begin{table}[t]
    \centering
    \resizebox{0.65\linewidth}{!}
    {
        \begin{tabular}{ccc|cc}
        \toprule
            ISF    &  ER  &  DSR   &   MLVU$_{dev}$    & VideoMME \\ \midrule
            & & & 60.9 & 57.1 \\
            \checkmark & & & 61.2 & 57.3 \\
            \checkmark & \checkmark & & 62.1 & 58.1\\
            \checkmark & \checkmark & \checkmark & 62.3 & 58.5 \\
        \hline
        \end{tabular}
    }
    \caption{Performance with different approaches for data exploration efficiency.}
    \vspace{-3mm}
    \label{tab:exploration}
\end{table}

\noindent
\textbf{Analysis of the data exploration efficiency}.
We investigate the impact of data exploration efficiency on model performance, as shown in Table~\ref{tab:exploration}. 
As discussed in Section 3.2, we employ three approaches to enhance data exploration efficiency: Ineffective Sample Filtering (ISF), Experience Replay (ER), and Dynamic Sample Ratio (DSR). 
We find that filtering out ineffective samples improves model performance, indicating that samples that are guessed correctly in blind tests are detrimental to the learning of the policy network. 
When implementing experience replay, the performance shows significant improvement as this allows each sample more exploration opportunities, leading to more stable training. 
The dynamic sample ratio mechanism also enhances model performance by enabling differentiated exploration for samples with varying difficulty levels.

\begin{table}[t]
    \centering
    \resizebox{0.6\linewidth}{!}
    {
        \begin{tabular}{c|cc}
        \toprule
            Strategy & MLVU$_{dev}$ & VideoMME  \\ \midrule
            FrameAvg     & 61.2 & 57.4  \\
            FrameAda     & 62.0 & 58.1  \\
            FrameAda+ST  & 62.3 & 58.5  \\
        \hline
        \end{tabular}
    }
    \caption{Performance with different token retention strategy.}
    \vspace{-5mm}
    \label{tab:retention}
\end{table}

\noindent
\textbf{Analysis of the token retention strategy}.
As shown in Table~\ref{tab:retention}, we study token retention strategies during inference, proposing three approaches: 
1) retaining equal tokens across all frames, FrameAvg. 
2) adaptively allocating token budgets per frame based on contribution scores of frames, FrameAda.
3) further enhancing spatio-temporal structural information by supplementing FrameAda with spatio-temporal tokens, FrameAda+ST. 
Comparing 1 and 2 reveals that allocating more tokens to frames with higher contributions improves model performance. 
Comparing 2 and 3 further shows that sacrificing a portion of sub-contributed tokens to maintain the video's spatiotemporal structure enhances the video understanding capabilities.

\noindent
\textbf{Visualization} analysis is shown in appendix.

\section{Conclusion}
In this paper, we present CaCoVID, the first reinforcement learning-based token compression framework for video understanding that optimizes a policy network to rank and prune video tokens by actively exploring optimal token combinations to correct predictions.
We propose a novel combinatorial policy optimization algorithm with online combination space sampling, which effectively narrows exploration complexity while accelerating policy convergence.
Extensive experiments on diverse video understanding benchmarks demonstrate the effectiveness of CaCoVID.

\bibliography{aaai2026}

\begin{appendices}
\section{Details of Training Pipeline}

\subsection{Optimization for Frame Contribution Estimation}
In the main manuscript, we provide a detailed explanation of how to optimize the policy model for token contribution estimation using the combinatorial policy optimization (CPO) algorithm. 
Here, we supplement the details regarding the optimization of frame contribution estimation using CPO. 
Given a frame sample ratio $r_f$, we sample $g_f$ groups of $r_f\times t$ frames $\{\mathcal{S}^{(i)}_f\}_{i=1}^{g_f}$ using online combinatorial space sampling strategy.
To ensure training efficiency for the joint optimization of tokens and frames, we employ the DPC-KNN algorithm to identify $r\times n_{vid}$ density peak tokens from the sampled frames as representatives for these frames, thereby aligning the number of sampled tokens for token and frame parallel optimization (where $r$ denotes the token sample ratio in the main manuscript).
We denote the density peak tokens of sampled frames as $\{\hat{X}_{f}^{(i)}\}_{i=1}^{g_f}$.
We feed $\{\hat{X}_{f}^{(i)}\}_{i=1}^{g_f}$ along with question tokens $X_{qst}$ into LLM for reasoning.
The correctness of the predictions is evaluated by comparing them with the answer $\mathcal{A}$ as the reward for the video frame combinations. Formally,
\begin{gather}
    X^{(i)}_f=[\hat{X}_{f}^{(i)},X_{qst}], \quad i = 1, 2, \dots, g_f\\
    {Y}^{(i)}_f = \text{LLM}(X^{(i)}_f), 
    R^{(i)}_f={\rm Reward}({Y}^{(i)}_f,\mathcal{A}).
\end{gather}
where the reward functions ${\rm Reward}(\cdot,\cdot)$ are consistent with Video-R1~\cite{video-r1}. Then, we compute the group advantage for policy optimization.
Formally,
\begin{gather}
    A^{(i)}_f = \frac{R^{(i)}_f-{\rm mean}(\{R^{(i)}_f\}_{i=1}^{g_f})}{{\rm std}(\{R^{(i)}_f\}_{i=1}^{g_f})}.
\end{gather}
Finally, we optimize the policy model for frame contribution estimation by maximizing the following objective,
\begin{gather}
    \mathcal{J}_{CPO}^f(\theta)=\mathbb{E}_{j\in\mathcal{Z}_f}\left[\frac{1}{g_f}\sum_{i=1}^{g_f}{r_{i,j}^{clip}(\theta)}\right],\\
    \text{\footnotesize $r_{i,j}^{clip}(\theta)={\rm min}\left[r_{i,j}(\theta)A^{(i)}_f,{\rm clip}\left(r_{i,j}(\theta),1-\epsilon_{low},1+\epsilon_{high}\right)A^{(i)}_f\right]$}, \\
    r_{i,j}(\theta)=\frac{\pi_{\theta}^f\left(j|[X_{vid},X_{qst}],\mathcal{S}_f^{(i)}\right)}{\pi_{\theta_{old}}^f\left(j|[X_{vid},X_{qst}],\mathcal{S}_f^{(i)}\right)}, \quad j\in \mathcal{Z}_f
\end{gather}
where $\mathcal{Z}_f=\{1, 2, \dots, t\}$ is the universal set of the video frame indexes, $\epsilon_{low},\epsilon_{high}$ is a clipping hyper-parameter~\cite{dapo} for stabilizing training, $\pi_{\theta}^f\left(\cdot|\cdot,\cdot\right)$ and $\pi_{\theta_{old}}^f\left(\cdot|\cdot,\cdot\right)$ denote the current and old compression policy network with parameters $\theta$ and $\theta_{old}$, whose output can be expressed as,
    \[
\pi_{\theta}^f\left(j|[X_{vid},X_{qst}],\mathcal{S}_f^{(i)}\right) = 
\begin{cases} 
\sigma(S_f)[j,1], & \text{if } j\in \mathcal{S}_f^{(i)} \\
\sigma(S_f)[j,0], & \text{if } j\in \mathcal{Z}_f\setminus\mathcal{S}_f^{(i)}
\end{cases}
\]
where $\sigma(\cdot)$ means softmax along the channel dimension.

\subsection{Training Pipeline}
To more clearly illustrate the training pipeline of CaCoVID, we provide a detailed algorithmic workflow in Algorithm~\ref{alg:algo}.
The token sample ratio is set as $r=0.02$.
The frame sample ratio is set as $r_f=0.125$.
$N_{data}$ is the dataset size.
$N_{iter}=5$ is the iteration times for each training sample.
$\alpha_{high}=0.875,\alpha_{low}=0.125$ are the thresholds for the dynamic sample ratio.
On one hand, our algorithm can narrow down the exploration space and leverage online learning experiences to minimize ineffective exploration through the online combinatorial space sampling strategy.
On the other hand, we employ experience replay with dynamic sample ratios during training to enhance the utilization efficiency of training samples, thereby stabilizing the training process and accelerating policy convergence.

\subsection{Operating Environment}
All experiments were conducted on eight H100 GPU with 80 GB memory and an Intel(R) Xeon(R) CPU, using a software environment consisting of \texttt{Python 3.10.16}, \texttt{PyTorch 2.5.1}, and \texttt{Transformers 4.50.0}.
All the experimental results are run three times to ensure reproducibility.

\begin{algorithm*}[t]
\caption{Training pipeline of CaCoVID.}
\small
\label{alg:algo}
\begin{algorithmic}[1]
\Require
    Dataset $\mathcal{D}$, policy network $\pi_\theta$, pretrained video LLM, token sample ratio $r$, frame sample ratio $r_f$;
\For {$k = 1,...,N_{data}$}
    \State Initialize dynamic token sample ratio $r^\prime\leftarrow r$;
    \State Initialize experience memory $\mathcal{M}\leftarrow\emptyset$;
    \State Set the old policy $\pi_{\theta_{old}}\leftarrow\pi_{\theta}$;
    \State Load the video-question pair: $\{V,Q\} \leftarrow \mathcal{D}[k]$;
    \State Encode video tokens: $X_{vid}\leftarrow {\rm VisonEncoder}(V)$;
    \State Embed question tokens: $X_{qst}\leftarrow {\rm TextEmbedding}(Q)$;
    \State Estimate the contribution of token and frames from the old policy: $S_t^{old},S_f^{old}={\pi_{\theta_{old}}}(X_{vid},X_{qst})$;
    \For {$j = 1,...,N_{iter}$}
        \State Sample indexs of tokens and frames by OCSS with ratio $r^\prime$ and $r_f$:
        \Statex \qquad \qquad $\{\mathcal{S}^{(i)}_t\}_{i=1}^{g_t}\leftarrow {\rm OCSS}(S_t^{old})$, $\{\mathcal{S}^{(i)}_f\}_{i=1}^{g_f}\leftarrow {\rm OCSS}(S_f^{old})$;
        \State Obtain sampled video tokens: $\{\hat{X}_t^{(i)}\}_{i=1}^{g_t},\{\hat{X}_f^{(i)}\}_{i=1}^{g_f}$;
        \State LLM reasoning with sampled video tokens:
        \Statex \qquad \qquad $\{{Y}^{(i)}_t\}_{i=1}^{g_t} = \left\{{\rm LLM}([\hat{X}_{t}^{(i)},X_{qst}])\right\}_{i=1}^{g_t}$, $\{{Y}^{(i)}_f\}_{i=1}^{g_f} = \left\{{\rm LLM}([\hat{X}_{f}^{(i)},X_{qst}])\right\}_{i=1}^{g_f}$;
        \State Compute the reward of predictions: 
        \Statex \qquad \qquad $\{R^{(i)}_t\}_{i=1}^{g_t}={\rm Reward}(\{{Y}^{(i)}_t\}_{i=1}^{g_t},\mathcal{A})$, $\{R^{(i)}_f\}_{i=1}^{g_f}={\rm Reward}(\{{Y}^{(i)}_f\}_{i=1}^{g_f},\mathcal{A})$;
        \State Experience replay: 
        \Statex \qquad \qquad $\{\mathcal{S}^{(i)}_t,R^{(i)}_t\}_{i=1}^{j\times g_t}\leftarrow{\rm Concat}\left(\{\mathcal{S}^{(i)}_t,R^{(i)}_t\}_{i=1}^{g_t},{{\rm Read}_t(\mathcal{M})}\right),\{\mathcal{S}^{(i)}_f,R^{(i)}_f\}_{i=1}^{j\times g_f}\leftarrow{\rm Concat}\left(\{\mathcal{S}^{(i)}_f,R^{(i)}_f\}_{i=1}^{g_f},{{\rm Read}_f(\mathcal{M})}\right)$;
        \State Compute the advantage of rewards:
        \Statex \qquad \qquad $\{A^{(i)}_t\}_{i=1}^{j\times g_t}={\rm Advantage}(\{R^{(i)}_t\}_{i=1}^{j\times g_t})$, $\{A^{(i)}_f\}_{i=1}^{j\times g_f}={\rm Advantage}(\{R^{(i)}_f\}_{i=1}^{j\times g_f})$;
        \State Optimize the policy network by maximize:
        \Statex \qquad \qquad $\mathcal{J}_{CPO}^t\left(\left\{\mathcal{S}^{(i)}_t,A^{(i)}_t\right\}_{i=1}^{j\times g_t},\theta,\theta_{old}\right)+\mathcal{J}_{CPO}^f\left(\left\{\mathcal{S}^{(i)}_f,A^{(i)}_f\right\}_{i=1}^{j\times g_f},\theta,\theta_{old}\right)$;
        \State Save experience to memory:
        \Statex \qquad \qquad $\mathcal{M}\leftarrow{\rm Append}(\mathcal{M},\{\mathcal{S}^{(i)}_t,R^{(i)}_t\}_{i=1}^{g_t},\{\mathcal{S}^{(i)}_f,R^{(i)}_f\}_{i=1}^{g_f})$
        \If {${\rm Mean}(\{R^{(i)}_t\}_{i=1}^{g_t})<\alpha_{low}$}
            \State $r^{\prime}\leftarrow r^{\prime}\times 2$;
        \ElsIf {${\rm Mean}(\{R^{(i)}_t\}_{i=1}^{g_t})>\alpha_{high}$}
            \State $r^{\prime}\leftarrow r^{\prime}/ 2$;
        \EndIf
    \EndFor
\EndFor
\end{algorithmic}
\end{algorithm*}

\section{Details of Comparison Methods}

All evaluation experiments are conducted using the LMMs-Eval\footnote{https://github.com/EvolvingLMMs-Lab/lmms-eval}~\cite{lmms} framework. 
We reimplement other compression algorithms within the same software environment and code framework for a fair comparison, following their official open-source implementations.

\noindent
Given the retention ratio of each layer $\{r_{i}\}_{i=1}^{N_{layer}}$, we define the equivalent retention ratio $r_{equ}$ as the layer-averaged retention ratio following FrameFusion~\cite{framefusion} to fairly compare compression methods applied in the LLM prefilling phase. Formally,
\begin{gather}
    r_{equ} = \frac{1}{N_{layer}}\sum_{i=1}^{N_{layer}}{r_{i}}.
\end{gather}

\noindent
\textbf{FastV}\footnote{https://github.com/pkunlp-icler/FastV} (ECCV2024).
FastV~\cite{fastv} retains video tokens with a ratio of $ r $ after the $ K^{th} $ LLM layer based on the attention scores from the LLM. 
We implement FastV with $ K=2 $.
$r$ is assigned to ensure the equivalent retention ratio aligns with the retention ratio presented in Table~1.

\noindent
\textbf{VisionZip}\footnote{https://github.com/dvlab-research/VisionZip} (CVPR2025).
We implement VisionZip~\cite{visionzip} after the pooling operation of visual features and prior to inputting them into the LLM. 
Each frame retains dominant and contextual tokens at a ratio of 54:10, following the original configuration. 
The overall retention ratio of video tokens aligns with the retention ratio in Table~1.

\noindent
\textbf{DivPrune}\footnote{https://github.com/vbdi/divprune} (CVPR2025).
We implement DivPrune~\cite{divprune} after the pooling operation of visual features and prior to inputting into the LLM. 
The ratio of video tokens retained matches the retention ratio in Table~1.

\noindent
\textbf{PruneVID}\footnote{https://github.com/visual-ai/prunevid} (ACL2025).
To directly compare the performance of compression before LLM prefilling, we implement spatial-temporal token merging of PruneVID~\cite{prunevid} after the pooling operation of visual features and prior to inputting into the LLM.
We set the temporal segment ratio $\gamma=0.25$, threshold $\tau=0.8$, token selection ratio $\alpha=0.4$, following the original configuration.
The cluster ratio $\beta$ controls the number of retained video tokens, set to $1.1\times$ the retention ratio.
This ensures the final number of retained video tokens closely aligns with the retention ratio in Table~1.

\noindent
\textbf{FrameFusion}\footnote{https://github.com/thu-nics/FrameFusion} (ICCV2025).
FrameFusion~\cite{framefusion} merges tokens whose inter-frame similarity is higher than the similarity threshold $S_{\rm threshold}$ in shallow layers to reduce redundant video tokens. 
When the number of similar tokens falls below a threshold $N_{\rm threshold}$, it further prunes unimportant tokens based on cumulative attention scores to achieve the target equivalent retention rate.
Following the original configuration, we set $S_{\rm threshold}=0.6$ and $N_{\rm threshold}=0.1$. 
The target equivalent retention rate aligns with the retention ratio in Table~1.
FrameFusion on Qwen2.5-VL-3B fails to achieve a 10\% equivalent retention ratio due to retaining an excessive number of tokens in the shallow layers.

\section{Extra Experimental Analysis}

\begin{table}[t]
    \centering
    \resizebox{\linewidth}{!}
    {
        \begin{tabular}{c|c|c|c}
        \toprule
            Method   & Compression Time (ms)$\downarrow$ & LLM Prefilling Time (ms)$\downarrow$ & Avg. Acc.$\uparrow$ \\ 
            \midrule
            Vanilla  & -     & 220.8 & 55.6 (100\%)  \\ \midrule
            DivPrune & 350.9 & 105.3 & 53.9 (97.0\%) \\ 
            PruneVID & 54.1  & 105.3 & 52.6 (94.6\%) \\\midrule
            CaCoVID  & 9.9   & 105.3 & 54.2 (97.5\%) \\
        \hline
        \end{tabular}
    }
    \caption{Comparison of compression efficiency on Qwen2.5-VL-3B with 25\% retention ratio.}
    \label{tab:efficiency}
\end{table}

\textbf{Analysis of the compression efficiency on Qwen2.5-VL}.
As shown in Table~\ref{tab:efficiency}, our algorithm achieves superior performance with only one-fifth of the latency required by other compression algorithms.
On one hand, the policy network in CaCoVID can estimate the contribution of each token to accurate predictions in parallel, thereby reducing compression latency. 
On the other hand, our contribution-aware token compression for video understanding explicitly aligns the objective of the compression policy with the correct prediction of LLM, thereby achieving superior performance.

\begin{table}[t]
    \centering
    \resizebox{0.75\linewidth}{!}
    {
        \begin{tabular}{c|c|cc}
        \toprule
            Inputs & Arch & MLVU$_{dev}$ & VideoMME  \\ \midrule
            $vid$      & 1$\times$Attn & 59.2 & 55.7  \\
            $vid+qst$  & 1$\times$Attn & 62.3 & 58.5  \\
            $vid+qst$  & 2$\times$Attn & 62.4 & 58.5  \\
        \hline
        \end{tabular}
    }
    \caption{Performance with different compression policy network on LLaVA-OneVision under 25\% retention ratio.}
    \label{tab:policy}
\end{table}

\noindent
\textbf{Analysis of the compression policy network}.
We explored the design of the policy network in terms of modality interaction and parameter scale. As shown in Table~\ref{tab:policy}, when we solely used video tokens as the inputs for token selection, the performance decreased significantly. 
When employing both video tokens and question tokens as the input of the policy network, the token compression policy achieves superior performance. 
This is because pure vision-based importance region estimation might prune critical tokens essential for correctly answering questions, which leads to performance degradation. 
Furthermore, increasing the attention network depth to two layers resulted in no substantial performance gains, indicating that a single-layer attention mechanism suffices to learn token selection capabilities correlated with question relevance. 
The additional computational cost introduced by deeper attention networks does not yield corresponding performance improvements. 
Therefore, we adopt a single-layer attention network combined with MLPs as the architecture for our compression policy network.

\begin{table}[t]
    \centering
    \resizebox{0.5\linewidth}{!}
    {
        \begin{tabular}{c|c|cc}
        \toprule
            $\lambda$ & MLVU$_{dev}$ & VideoMME  \\ \midrule
            $1/r$ & 59.2 & 55.1  \\
            4.0   & 61.7 & 58.1  \\
            2.0   & 62.3 & 58.5  \\
            1.0   & 61.9 & 57.7  \\
        \hline
        \end{tabular}
    }
    \caption{Performance with different size of combinatorial sub-space on LLaVA-OneVision under 25\% retention ratio.}
    \label{tab:lambda}
\end{table}

\noindent
\textbf{Analysis of the combinatorial sub-space division}.
Given a token sampling ratio $ r $, we sort all tokens based on their estimated contribution scores of the policy network, and then divide them into $ l=1/(\lambda\times r) $ combinatorial sub-spaces, each containing $ m=\lambda \times r \times n_{vid} $ tokens for OCSS.
We set $\lambda=2$ by default in the main manuscript.
Here, we further investigate the impact of the size of the combinatorial sub-space controlled by $\lambda$ on compression performance.
As shown in Table~\ref{tab:lambda}, the compression policy performs best with $\lambda=2$.
Larger combinatorial sub-spaces (larger $\lambda$) are more likely to explore optimal token combinations, but they also introduce a large number of ineffective combinations containing both high-contribution tokens and low-contribution tokens, which hinder model optimization.
The results in Table~\ref{tab:lambda} demonstrate that $\lambda=2$ strikes a good balance and achieves the best performance.

\section{Visualization}

\begin{figure}[t]
    \centering
    \vspace{-2mm}
    \includegraphics[width=1.0\linewidth]{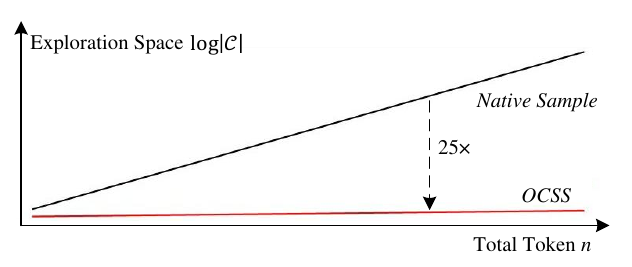}
    \caption{Our OCSS can reduce the logarithmic exploration space complexity ${\rm log(\mathcal{C})}$ to 1/25 of arbitrary exploration magnitude.}
    \vspace{-2mm}
    \label{fig:sample_space}
\end{figure}

\begin{figure*}[tp]
    \centering
    \vspace{-2mm}
    \includegraphics[width=1.0\linewidth]{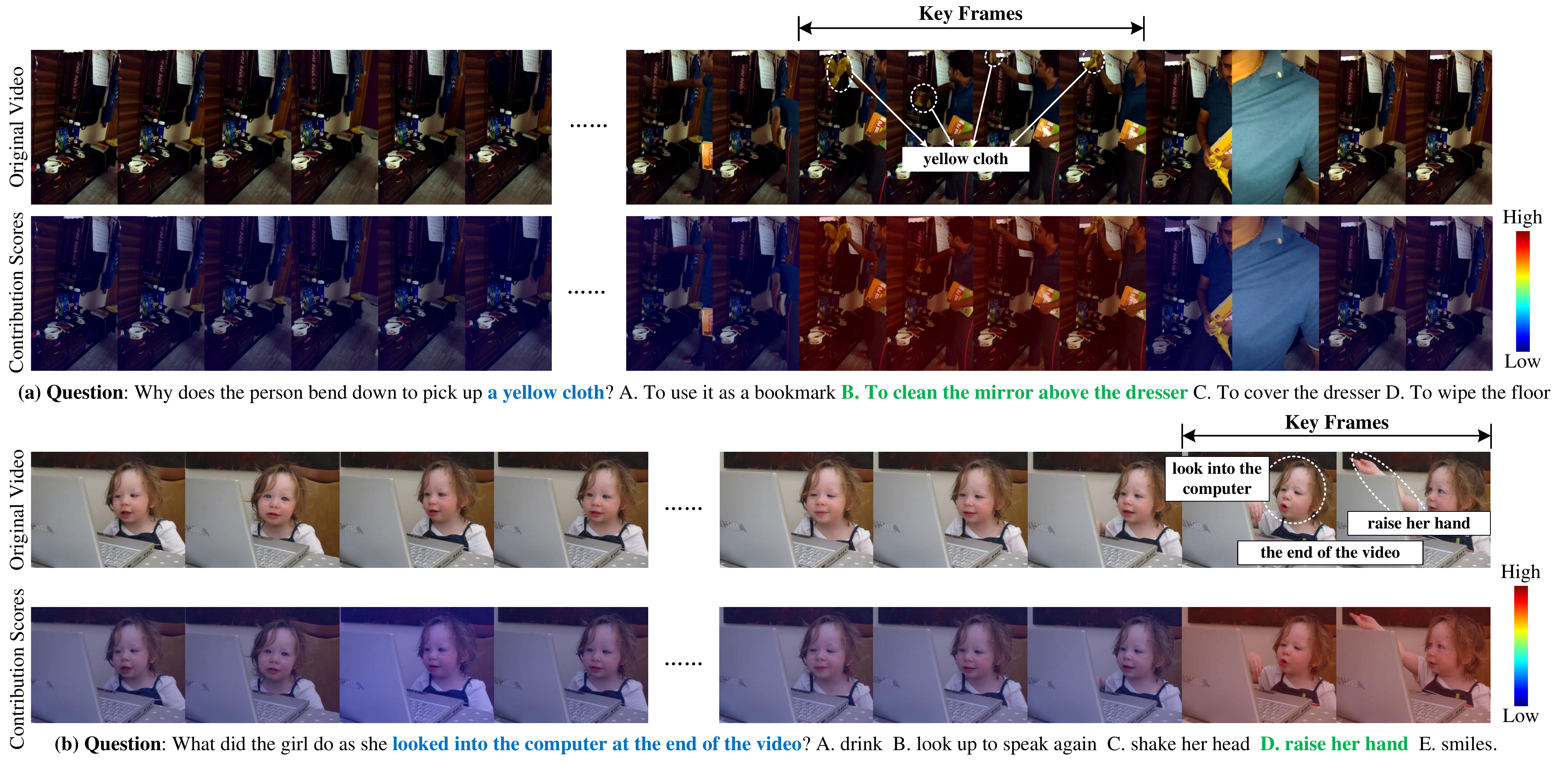}
    \caption{The compression policy network can effectively identify the most critical frames to answer the question, such as the frames when the man picks up a yellow cloth and the frames when the girl looks into the computer at the end of the video.}
    \vspace{-3mm}
    \label{fig:frame_case}
\end{figure*}

\begin{figure}[tp]
    \centering
    \includegraphics[width=1.0\linewidth]{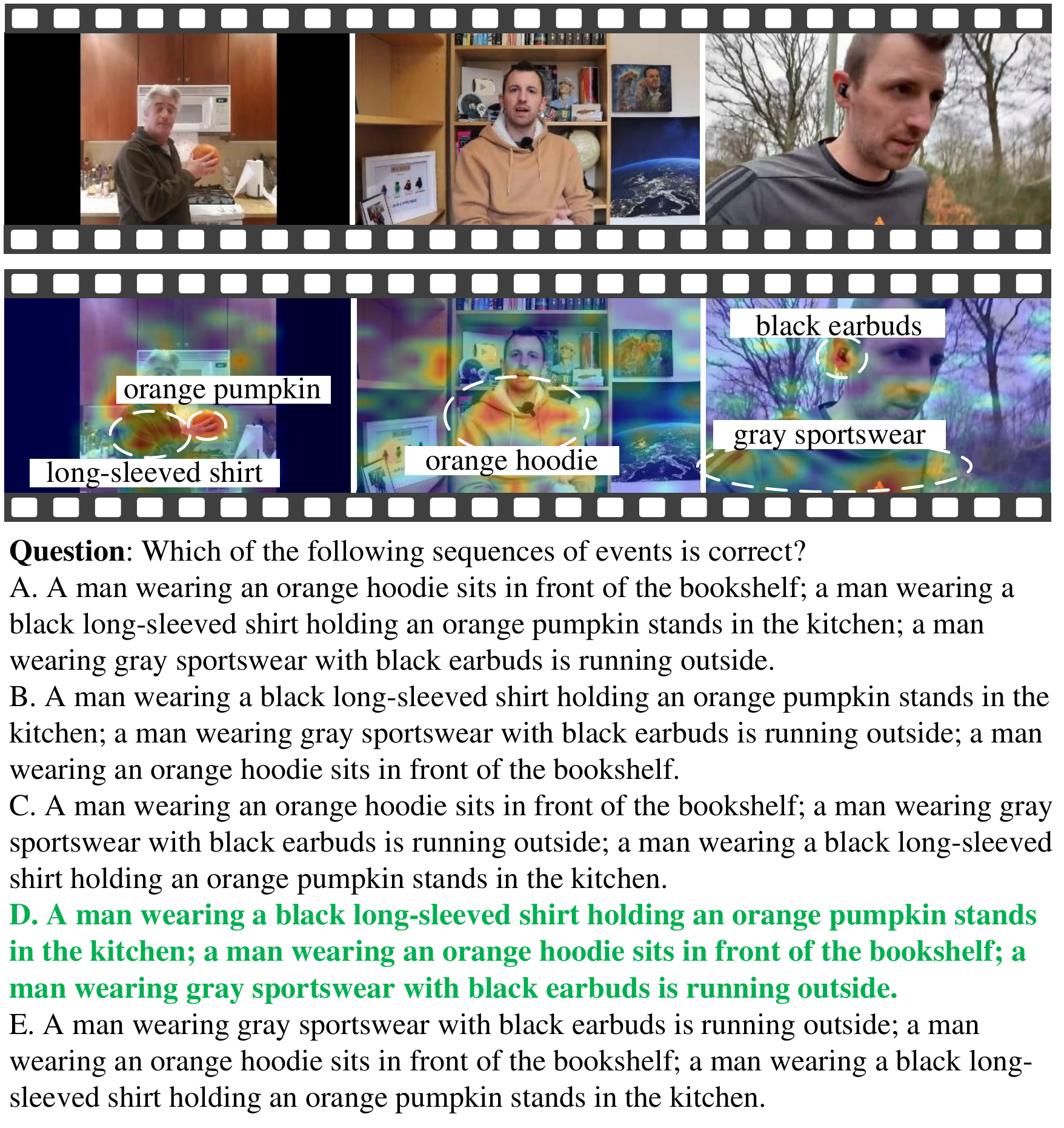}
    \caption{
    Our compression policy network can handle complex video understanding and identify the most critical tokens relevant to the question, such as orange pumpkin, long-sleeved shirt, orange hoodie and black earbuds.}
    \label{fig:vis_complex}
\end{figure}
\begin{figure*}[t]
    \centering
    \includegraphics[width=0.9\linewidth]{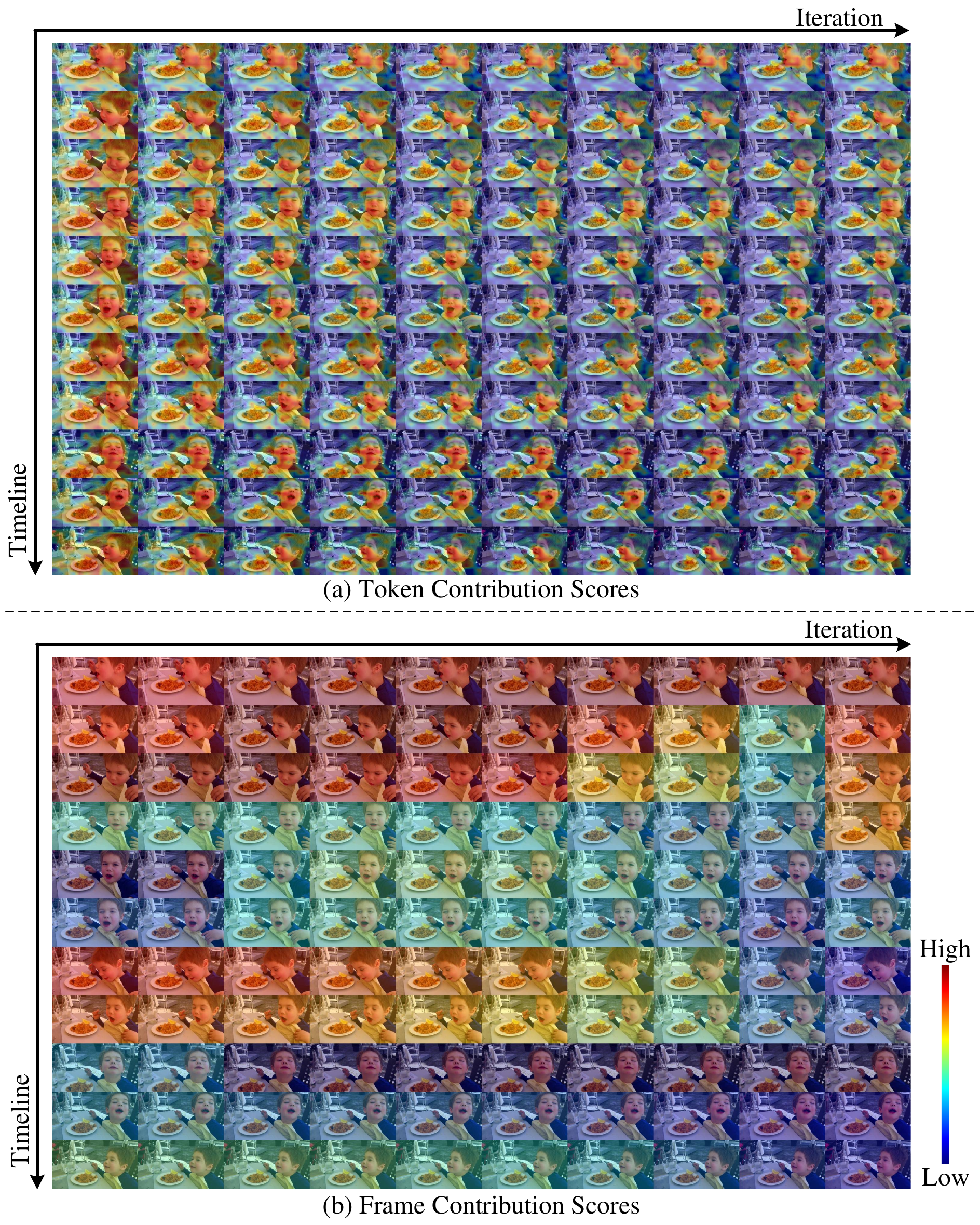}
    \caption{The question of the video is ``Why was the boy coughing at the start?" and the answer is ``choke". With our combinatorial policy optimization, the compression policy network with a limited number of iterations can attend to the most critical tokens for accurate responses: food and mouths, and the most critical frames for accurate responses: the beginning frames of the video. This visually demonstrates the effectiveness of our combinatorial policy optimization for token and frame contribution estimation.}
    \label{fig:optimization_our}
\end{figure*}
\begin{figure*}[t]
    \centering
    \includegraphics[width=0.9\linewidth]{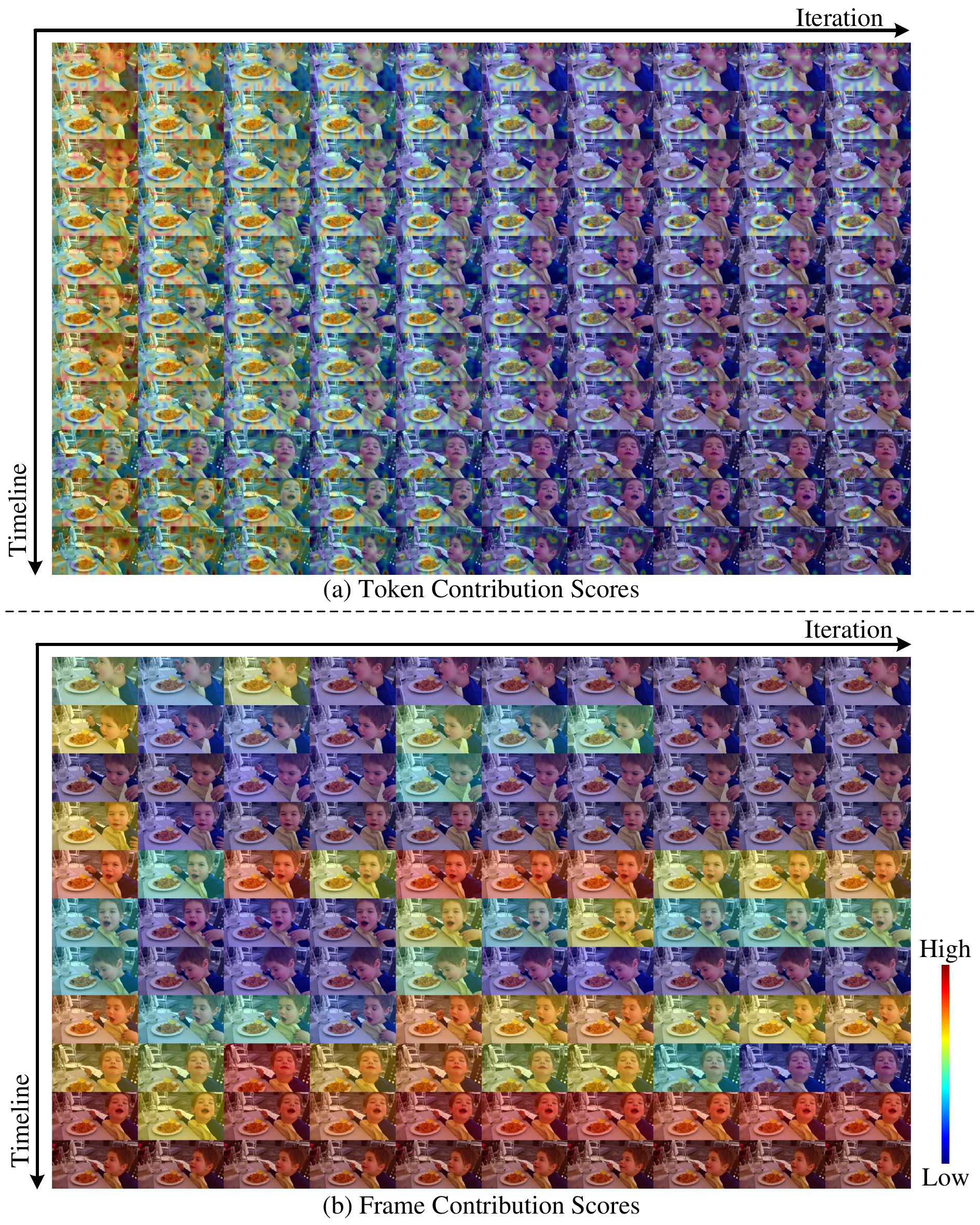}
    \caption{The question of the video is ``Why was the boy coughing at the start?" and the answer is ``choke". Without online combinatorial space sampling strategy, the compression policy network is prone to divergent learning dynamics. This is because the search space of token combinations is extremely vast, making it challenging to discover optimal token combinations for accurate predictions. Furthermore, extensive sampled combinations often contains both high-contribution tokens and low-contribution tokens, which may mislead the policy network towards suboptimal optimization directions. Differently, our online combinatorial space sampling strategy effectively addresses these challenges, enabling stable policy network optimization.}
    \label{fig:optimization_ori}
\end{figure*}

\noindent
\textbf{Visualization of combinatorial policy optimization}.
In combinatorial policy optimization, we employ an online combinatorial space sampling strategy (OCSS) to explore the optimal token combinations. 
OCSS employs a two-stage sampling mechanism by combinatorial sub-space division to effectively reduce the exploration space of token combinations.
As illustrated in Figure~\ref{fig:sample_space}, compared to randomly exploring arbitrary token combinations ${\rm O}(2^n)$, our OCSS algorithm reduces the logarithmic exploration space complexity $\log(\mathcal{C})$ by a factor of 25.
Furthermore, by partitioning the combinatorial sub-space, OCSS ensures that sampled tokens possess similar contributions rather than indiscriminately exploring arbitrary combinations. 
This enables the policy network to systematically explore token combinations with the highest potential to guide the LLM toward correct answers for given questions.
As illustrated in Figure~\ref{fig:optimization_our}, with OCSS, the compression policy network can focus on the most critical tokens ($e.g.$, food and mouth) and frames ($e.g.$, the beginning of the video) within fewer than 10 iterations to answer the question.
As illustrated in Figure~\ref{fig:optimization_ori}, without OCSS, the compression policy network is prone to divergent learning dynamics. This is because the search space of token combinations is extremely vast, making it challenging to discover optimal token combinations for accurate predictions. Furthermore, extensive sampled combinations often contain both high-contribution tokens and low-contribution tokens, which may mislead the policy network toward suboptimal optimization directions.
These visualization results demonstrate the effectiveness of our combinatorial policy optimization algorithm.

\noindent
\textbf{Visualization of contribution scores of frames}.
We provide the visualization of frame contribution scores for video understanding in Figure~\ref{fig:frame_case}.
Our compression policy network can effectively focus on question-critical frames, such as the frames when the man picks up a yellow cloth and the frames when the girl looks into the computer at the end of the video.
These visualization results provide intuitive evidence that the trained compression policy network can effectively identify the critical frames essential for correctly answering the questions.

\noindent
\textbf{Visualization of contribution scores of tokens}.
We further provide the visualization of token contribution scores for complex video understanding in Figure~\ref{fig:vis_complex}.
Our compression policy network can effectively focus on question-critical regions such as clothing, pumpkin and earbud.
This provides intuitive evidence of the generalization capability of our compression policy network in complex video understanding tasks.

\end{appendices}

\end{document}